\documentclass[sigconf]{acmart}

\usepackage{amsmath,amsfonts,bm}

\def\eqref#1{equation~\ref{#1}}

\def\1{\bm{1}}

\DeclareMathAlphabet{\mathsfit}{\encodingdefault}{\sfdefault}{m}{sl}
\SetMathAlphabet{\mathsfit}{bold}{\encodingdefault}{\sfdefault}{bx}{n}

\usepackage{hyperref}
\usepackage{amsmath}
\usepackage{url}
\usepackage{graphicx}
\usepackage{longtable}
\usepackage{listings}
\usepackage{color}
\usepackage{enumitem}
\usepackage{tikz}
\usepackage[most]{tcolorbox}

\definecolor{dkgreen}{rgb}{0,0.6,0}
\definecolor{gray}{rgb}{0.5,0.5,0.5}
\definecolor{mauve}{rgb}{0.58,0,0.82}

\AtBeginDocument{%
  \providecommand\BibTeX{{%
    \normalfont B\kern-0.5em{\scshape i\kern-0.25em b}\kern-0.8em\TeX}}}

\copyrightyear{2023}
\acmYear{2023}
\setcopyright{rightsretained}
\acmConference[KDD '23]{Proceedings of the 29th ACM SIGKDD Conference on Knowledge Discovery and Data Mining}{August 6--10, 2023}{Long Beach, CA, USA} \acmBooktitle{Proceedings of the 29th ACM SIGKDD Conference on Knowledge Discovery and Data Mining (KDD '23), August 6--10, 2023, Long Beach, CA, USA} \acmDOI{10.1145/3580305.3599827} \acmISBN{979-8-4007-0103-0/23/08}

\begin{document}

\title{From Human Days to Machine Seconds:\\Automatically Answering and Generating\\Machine Learning Final Exams}


\author{Iddo Drori}
\affiliation{%
  \institution{MIT, Columbia University, BU}
  \city{Cambridge}
  \country{USA}
}
\email{idrori@mit.edu}

\author{Sarah J. Zhang}
\affiliation{%
  \institution{MIT}
  \city{Cambridge}
  \country{USA}}
\email{sjzhang@mit.edu}

\author{Reece Shuttleworth}
\affiliation{%
  \institution{MIT}
  \city{Cambridge}
  \country{USA}
}
\email{rshuttle@mit.edu}

\author{Sarah Zhang}
\affiliation{%
 \institution{MIT}
 \city{Cambridge}
 \country{USA}
}
\email{sazhang@mit.edu}

\author{Keith Tyser}
\affiliation{%
  \institution{Boston University}
  \city{Boston}
  \country{USA}
}
\email{ktyser@bu.edu}

\author{Zad Chin}
\affiliation{%
  \institution{Harvard University}
  \city{Cambridge}
  \country{USA}
}
\email{zadchin@college.harvard.edu}

\author{Pedro Lantigua}
\affiliation{%
  \institution{MIT}
  \city{Cambridge}
  \country{USA}}
\email{lantigua@mit.edu}

\author{Saisamrit Surbehera}
\affiliation{%
  \institution{Columbia University}
  \city{New York}
  \country{USA}}
\email{ss6365@columbia.edu}

\author{Gregory Hunter}
\affiliation{%
  \institution{Columbia University}
  \city{New York}
  \country{USA}}
\email{geh2129@columbia.edu}

\author{Derek Austin}
\affiliation{%
  \institution{Columbia University}
  \city{New York}
  \country{USA}}
\email{da2986@columbia.edu}

\author{Leonard Tang}
\affiliation{%
  \institution{Harvard University}
  \city{Cambridge}
  \country{USA}}
\email{leonardtang@college.harvard.edu}

\author{Yann Hicke}
\affiliation{%
  \institution{Cornell University}
  \city{Ithaca}
  \country{USA}}
\email{ylh8@cornell.edu}

\author{Sage Simhon}
\affiliation{%
  \institution{MIT}
  \city{Cambridge}
  \country{USA}}
\email{simhon@mit.edu}

\author{Sathwik Karnik}
\affiliation{%
  \institution{MIT}
  \city{Cambridge}
  \country{USA}}
\email{skarnik@mit.edu}

\author{Darnell Granberry}
\affiliation{%
  \institution{MIT}
  \city{Cambridge}
  \country{USA}}
\email{darnellg@mit.edu}

\author{Madeleine Udell}
\affiliation{%
  \institution{Stanford University}
  \city{Stanford}
  \country{USA}}
\email{udell@stanford.edu}
\renewcommand{\shortauthors}{Drori et al.} 

\definecolor{purple}{rgb}{0.3,0.0,.4}

\begin{abstract}
A final exam in machine learning at a top institution such as MIT, Harvard, or Cornell typically takes faculty days to write, and students hours to solve. We demonstrate that large language models pass machine learning finals at a human level, on finals available online after the models were trained, and automatically generate new human-quality final exam questions in seconds. Previous work has developed program synthesis and few-shot learning methods to solve university-level problem set questions in mathematics and STEM courses. In this work, we develop and compare methods that solve final exams, which differ from problem sets in several ways: the questions are longer, have multiple parts, are more complicated, and span a broader set of topics. We curate a dataset and benchmark of questions from machine learning final exams available online and code for answering these questions and generating new questions. We show how to generate new questions from other questions and course notes. For reproducibility and future research on this final exam benchmark, we use automatic checkers for multiple-choice, numeric, and questions with expression answers. We perform ablation studies comparing zero-shot learning with few-shot learning and chain-of-thought prompting using GPT-3, OPT, Codex, and ChatGPT across machine learning topics and find that few-shot learning methods perform best. We highlight the transformative potential of language models to streamline the writing and solution of large-scale assessments, significantly reducing the workload from human days to mere machine seconds. Our results suggest that rather than banning large language models such as ChatGPT in class, instructors should teach students to harness them by asking students meta-questions about correctness, completeness, and originality of the responses generated, encouraging critical thinking in academic studies.
\end{abstract}


\maketitle

\section{Introduction}
\label{sec:introduction}
Can a machine learn machine learning and help teach it? We automatically solve, explain, and generate new questions in machine learning courses, reducing the efforts of instructors and TAs from days to seconds. We evaluate students on meta-questions that validate these models' correctness, completeness, and originality.

This work presents a dataset of machine learning final exams with 646 question parts and a benchmark of baselines using a variety of language models with different prompting schemes. The best of these performs at a human level. In university-level STEM courses, students complete assignments, including problem sets and labs, and exams throughout the course. Recent work has developed accurate methods to solve STEM problem sets \citep{drori2022math} using language models and few-shot learning. However, final exams remain challenging: final exams test the cumulative understanding of material learned over a semester and evaluate the students' depth and breadth of expertise. This work is the first to present a structured dataset of machine learning finals and a benchmark of baseline methods for answering them. 

Final exams differ from problem sets in several ways, and the experience of solving each varies. First, finals are long, containing around nine questions with around seven parts each. Final exam questions are also multifaceted and multi-stepped: different parts of a single question require applying different concepts and problem-solving skills, and parts may build upon each other. While weekly problem sets focus on a single topic, finals span topics from the entire semester. Further, final questions are often story-based problems that may require mathematical modeling. Due to the time constraint of these exams, finals are also designed to test core understanding and application of course material over rote calculations or hand computations, which are more often found on problem sets that help students build up a theoretical intuition for that week's material. Thus, asking a machine to answer questions from finals allows for testing whether the model is able to learn a breadth and depth of topics beyond problem sets. 

One contribution of this work is the dataset of final questions derived from three courses: Introduction to Machine Learning at MIT and Cornell and Machine Learning at Harvard. These are undergraduate courses with hundreds of students each semester, making them the largest undergraduate courses offered. 
At MIT, Introduction to Machine Learning is a core class in the computer science program. The prerequisites for the course are Linear Algebra and Fundamentals of Programming or Introduction to Algorithms. The class typically consists of weekly exercises, labs, quizzes, homework, a midterm, and a final exam.
The final exam questions have many parts, each posing a new problem, 
and each question in the dataset corresponds to one part. 
The questions test a variety of topics and require different solution types. 
Solutions are primarily open-ended questions with some true/false and multiple-choice questions on theory, math, and code implementations. Due to the diversity of the final questions, our dataset uniquely assesses advanced problem-solving and reasoning skills in machine learning, math, and natural language processing. 

We propose several baseline methods that solve these problems, by zero-shot and few-shot learning using GPT-3, Codex, OPT, and ChatGPT, and adding chain-of-thought prompting. We find that few-shot learning methods perform best. As shown in Table \ref{tab:man-vs-machine}, the best-performing methods pass the final exams, and their grade is comparable with human grades of students on the same machine learning finals evaluated by the same human graders. 
We verify that our results are not due to overfitting by testing the method on three finals that are available online after the model was trained.
We generate new final exam questions indistinguishable from human-written questions from other questions and course notes. 

\begin{table*}[h!]
\small
    \caption{Human and machine grading of human and machine solved final exams. Mean human and machine grades (out of 100) on MIT Introduction to Machine Learning final exams by semester. Non-image grades consider question parts that do not contain images that are required for solving the question.}
    \label{tab:man-vs-machine}
    \centering
    \begin{tabular}{lcccc}
        \toprule
        Grading: & Human & Human & Human & Machine\\
        Answers: & Human & Human & Machine & Machine\\
        Questions: & All & Non-Image & Non-Image & Non-Image Non-Open\\
        \midrule
        MIT Spring 2021 & 75.84 & 80.77 & 62.09 & 64 \\
        MIT Fall 2021 & 74.38 & 60.88 & 58.94 & 51.33 \\
        MIT Spring 2022 & 69.07 & 70.82 & 68.86 & 73.53 \\
        \midrule
        Mean & 73.10 & 70.82 & 63.29 & 62.95\\
        \bottomrule
    \end{tabular}
\end{table*}

In summary, the key contributions of our work are:
\begin{enumerate}
\item A new dataset of machine learning final exams.
\item A benchmark and baseline for answering final exam questions.
\item A comparison of different methods and their solve rates.
\item The automatic grading of checkable question types forming a usable benchmark.
\item The generation of new questions from other questions.
\item The generation of new questions from course notes.
\item The development of meta-questions about correctness, completeness, and originality.
\item A survey showing generated questions are indistinguishable from human-written questions.
\end{enumerate}

\subsection{Related Work}
It is often thought that humans are generalists, whereas machines are specialists. However, large language models such as GPT-3 \citep{brown2020language}, Gopher \citep{rae2021scaling}, PaLM \citep{chowdhery2022palm}, BLOOM \citep{scao2022bloom}, and ChatGPT \citep{openai2022chatgpt}, also called foundation models, are generalist learners. Specifically, in our setting, while humans care about the number of topics in an exam and therefore find finals more difficult than problem sets, foundation models effortlessly scale to many topics without re-training. Language models may be pre-trained on text and fine-tuned on specific datasets, for example OpenAI's Codex \citep{chen2021evaluating} and ChatGPT \citep{openai2022chatgpt} allow for generating programs and answering questions from text at a human level. There are several ways to improve the mathematical reasoning ability of language models: (1) using chain-of-thought (CoT) prompting \citep{kojima2022large, wei2022chain}, (2) using the top-k ranking solutions \citep{li2022competition} and merging them by voting \citep{wang2022self} or least-to-most prompting \citep{zhou2022least}, and (3) using program synthesis and few-shot learning to generate code that answers questions \citep{drori2022math}.

Much of the prior work focuses on high school or middle school level material \citep{qu-etal-2021-asking}. The first work to tackle university-level machine learning course problem set questions \citep{tran2021solving} used a transformer and GNN architecture and heavily relied on data augmentation. This resulted in overfitting and did not scale up to other types of questions or courses. Probability and statistics course problem-set questions have been answered \citep{tang2022stats} by probabilistic program synthesis with human performance. Problem-set questions from the core university math courses \citep{drori2022math} have been automatically solved using few-shot learning and program synthesis at a human level. Other work considers university-level course questions across a variety of domains \citep{hendrycks2020measuring} and identifying theorems \citep{srivastava2022beyond}. 
Prior work on question generation includes question--answer pair generation based on a text passage \citep{qu-etal-2021-asking} and question text generation based on other questions \citep{drori2022math}.

\section{Dataset}
\label{sec:dataset}

\begin{table}[tb!]
\small
    \centering
    \caption{The number of questions and parts in the final for each semester of machine learning courses.}
    \label{tab:exam-stats-by-semester}
    \begin{tabular}{lcc}
        \toprule
        Semester & Questions & Parts\\
        \midrule
        MIT Fall 2017   & 10 & 61\\
        MIT Spring 2018 & 9  & 42\\
        MIT Fall 2018   & 10  & 60\\
        MIT Spring 2019 & 9  & 58\\
        MIT Fall 2019   & 8  & 61\\
        MIT Spring 2021 & 13 & 71\\
        MIT Fall 2021   & 8  & 86\\
        MIT Spring 2022 & 9  & 59\\
        Harvard Spring 2015 & 8 & 12\\
        Harvard Spring 2021 & 6 & 32\\
        Cornell Spring 2017 & 30 & 48\\
        Cornell Fall 2018 & 29 & 56\\
        \midrule
        Mean & 12.42 & 53.83\\
        Total & 149 & 646\\
        \bottomrule
    \end{tabular}
    \vspace{-10pt}
\end{table}

\begin{table}[tb!]
\small
    \centering
    \caption{The number of questions and parts in all finals for each topic in the machine learning courses. Topics can have half-questions attributed to them if a question has some parts under one topic and the other parts under another topic.}
    \label{tab:exam-stats-by-topic}
    \begin{tabular}{lcccc}
        \toprule
        Topic & Questions & Parts\\
        \midrule
        Regression & 10 & 62\\
        Classifiers & 24 & 85\\
        Logistic Regression & 3 & 10\\
        Features & 3.5 & 21\\
        Neural Networks & 19.5 & 87\\
        Loss Functions & 4 & 16\\
        CNNs & 9 & 65\\
        MDPs & 10 & 77\\
        RNNs & 7 & 33\\
        Reinforcement Learning & 11 & 60\\
        Clustering & 5 & 17\\
        Decision Trees & 14 & 48\\
        Model Selection & 5 & 16\\
        Ensemble Methods & 9 & 20\\
        Bayesian Networks & 1 & 6\\
        HMMs & 1 & 4\\
        Optimization & 10 & 16\\
        Bonus & 3 & 3\\
        \midrule
        Mean & 8.28	& 35.89\\
        Total & 149 & 646\\
        \bottomrule
    \end{tabular}
    \vspace{-10pt}
\end{table}

We present a new dataset of 646 question parts from a dozen final exams of MIT's and Cornell's Introduction to Machine Learning courses and Harvard's Machine Learning class. Our dataset covers the finals given at MIT for semesters of Fall 2017, Spring 2018, Fall 2018, Spring 2019, Fall 2019, Spring 2021, Fall 2021, and Spring 2022, Harvard Spring 2015 and Spring 2021, and Cornell Spring 2017 and Fall 2018. Due to the COVID-19 pandemic, no finals were in the MIT course during 2020. The dataset spans questions on the 17 machine learning topics covered in the courses: (1) regression, (2) classifiers, (3) logistic regression, (4) features, (5) loss functions, (6) neural networks, (7) convolutional neural networks (CNNs), (8) Markov decision processes (MDPs), (9) recurrent neural networks (RNNs), (10) reinforcement learning, (11) clustering, (12) decision trees, (13) model selection, (14) ensemble methods, (15) Bayesian networks, (16) hidden Markov models (HMMs), and (17) optimization.

The breakdown of questions, parts, points, and non-image points by each semester and topic are shown in Tables \ref{tab:exam-stats-by-semester} and \ref{tab:exam-stats-by-topic}. Each question in a final exam consists of multiple parts. Questions are written by providing set-up and context information first, followed by the question parts (which may come with additional information). Set-up and context information may contain (1) story elements (ex., character names and motivations), (2) relevant definitions and equations, and (3) data points. We format questions in the dataset by concatenating the question context, any context or solutions from prior parts of the question required for answering the part, and the part's context and question. We split the questions into their corresponding parts. Questions consist of text, mathematical notation, and images. Mathematical notation is represented in the dataset by \LaTeX \space and images by screenshots from PDF documents. The types of question answers are diverse. A few are multiple-choice or true/false questions. Most are open-ended, for which the evaluation requires modeling the problem, mathematical manipulation, or writing code. Many questions require explaining the answer.

We use twelve final exams from different semesters for data curation. The PDF files of the exams are publicly available online. We use a tool \citep{MathpixSnip} for an initial transcription, and curators then evaluate and manually correct the input questions, and verify the correctness of each input question.

We extract questions and solutions for all parts of all types of questions, including those that rely on images. We curate nine exams from publicly available PDF documents. The three MIT exams between 2021 and 2022 are after the model training, therefore the model does not overfit their solutions. The aggregate average grades are available to the students and do not contain any personally identifiable information. There are three duplicate questions between exams which we keep.

\section{Benchmark}

\subsection{Baselines}
We provide a benchmark by comparing nine baselines for answering the final exam questions: (1) GPT-3 text-davinci-002 with zero-shot learning, (2) GPT-3 text-davinci-003 with zero-shot learning, (3) ChatGPT with zero-shot learning, (4) GPT-3 with few-shot learning, (5) GPT-3 with zero-shot learning and chain-of-thought (CoT) prompting, (6) GPT-3 with few-shot learning and chain-of-thought (CoT) prompting, (7) Codex with zero-shot learning, (8) Codex with few-shot learning, and (9) OPT with zero-shot learning.

GPT-3 zero-shot uses the question as-is, whereas GPT-3 zero-shot with CoT uses the suffix ``Let's think step by step.'' after the question to encourage multi-step output. Codex zero-shot uses the prefix ``Write a program that answers'' before the question within Python comments denoted by triple quotes """ to encourage Codex to write code. GPT-3 few-shot finds the closest questions in the embedding space, measured by cosine similarity, and uses them and their corresponding answers before the new question as examples in the prompt. Codex few-shot finds the closest questions in the embedding space, as measured by cosine similarity, and uses these questions and their corresponding code as examples. 

For students, a good study technique is to use previous final exams to review and practice for their upcoming final. We model this method by few-shot learning using the question--answer pairs (for GPT-3) or question--code (for Codex) with the closest question embeddings from previous finals. We implement this by considering all the exam questions, marking each question by its semester and year, and using only previous semesters' questions for few-shot learning. The MIT Fall 2017 and Spring 2022 exams contain three duplicate questions, and we handle these same questions the same way humans do by allowing few-shot learning in MIT Spring 2022 based on successful Fall 2017 zero-shot answers. It is reasonable that if a student studies all previous exams, there may be 8.5\% of repeated question points. Since MIT Fall 2017, Harvard Spring 2015, and Cornell Spring 2017 are the first final exams in the corresponding universities, and we do not perform few-shot learning on these. 

\subsubsection{Comparison with Open Language Models}
We also evaluate our dataset on an open-source language model, Meta's OPT-175B. OPT-175B is a model consisting of 175 billion parameters. Tables \ref{tab:performance-by-semester} and \ref{tab:performance-by-topic} compare the results of GPT-3, ChatGPT, Codex, and Meta OPT. We evaluated OPT on only 163 question parts since OPT was limited to handling questions under 256 characters in length. We implement the inference for the OPT-175B model using Alpa. Alpa is a particular framework designed for training and inference of large models. For the hardware, we use an 8x A100 PCIE cluster. The model requires about 560 GB of VRAM in our run case, and each example takes nine minutes for inference.

\subsection{Grading}

\subsubsection{Human Grading}
The questions are of different types: multiple-choice, numerical, expressions, and open-text. We grade answers and aim to keep all factors equal in grading human and machine answers. Human and machine answers are graded based on the number of points allocated to each question part, giving complete, partial, or no credit for each answer. We approximate partial credit by assigning half credit. The course staff graded student final exams, which included graduate TAs and instructors. The same graduate TAs and the instructor that graded the student answers also graded the machine answers. Grading instructions are the same for grading student answers as for machine answers.

\subsubsection{Automatic Grading}
We label each question's answer type into one or two categories out of four options - multiple-choice (MC), numerical, expression, or open. We consider answers multiple-choice if the test-taker is presented with an enumerated list of choices, numerical if the answer is a number, expression if the answer includes variables or other notation, and open if the answer calls for free-response text. We categorize questions that have additional questions nested within them by the multiple relevant categories. This is often the case when a question with one of MC, numerical, or expression is followed by a follow-up question asking the student to explain their previous answer.
The breakdown of the questions is as follows: 98 are multiple-choice, 84 are numerical, 81 are expressions, and 204 are open. The 'Non-Open Points' column of Tables \ref{tab:image-parts-by-semester} and \ref{tab:image-parts-by-topic} show the answer type breakdown by the number of points. Table \ref{tab:image-parts-by-semester} shows the number of question parts that do not rely on images, the number of points that do not rely on images and the number of non-open question points in Introduction to Machine Learning finals for each semester. Table \ref{tab:image-parts-by-topic} shows the breakdown by topic. Our automatic grading uses string matching and regular expressions. In the case of multiple-choice results, we check that the output of the code is equal to the solution. In the case of numerical answers, we look for a matching integer or real number.

\subsection{Performance}
Table \ref{tab:performance-by-semester} shows the machine grades by semester, and Table \ref{tab:performance-by-topic} shows the machine grades by topic, excluding question parts that rely on images. We compare the average grade (out of 100) of GPT-3 with zero-shot (ZS), GPT-3 with ZS and chain-of-thought (CoT) prompting, GPT-3 with few-shot (FS) learning, GPT-3 with FS and CoT prompting, Codex with ZS, Codex with FS, ChatGPT, and OPT with ZS. Fall 2017 is the first semester, so few-shot learning results based on previous semesters are unavailable (NA). Spring 2020 and Fall 2020 did not have final exams due to COVID-19. Three final exams were available online after GPT-3, ChatGPT, and Codex were trained, ensuring that the model is not overfitting content it has seen previously. The results consistently demonstrate that for zero-shot, ChatGPT is best and that few-shot learning methods perform best across semesters and topics, as marked in bold.

\begin{table*}[h!]
\small
    \caption{We benchmark different baselines for each semester, excluding question parts that rely on images. We compare the average grade of GPT-3 (text-davinci-002 and 003) with zero-shot (ZS), ChatGPT, GPT-3 with few-shot (FS) learning, GPT-3 with ZS and chain-of-thought (CoT) prompting, GPT-3 with FS and CoT prompting, Codex with ZS, Codex with FS, and OPT with ZS. MIT Fall 2017, Cornell Spring 2017, and Harvard Spring 2015 were the first semester for each university, so few-shot learning results based on previous semesters are unavailable (NA). The result of the best-performing method for each semester is marked in bold.}
    \label{tab:performance-by-semester}
    \centering
    \begin{tabular}{lccccccccc}
        \toprule
        Semester & GPT-3 2 ZS & GPT-3 3 ZS & ChatGPT ZS & GPT-3 FS & GPT-3 ZS CoT & GPT-3 FS CoT & Codex ZS & Codex FS & OPT ZS\\
        \midrule
        MIT Fall 2017 &	38.21 & \textbf{50.00} & 48.93 & NA & 22.86 & NA & 21.43 & NA & NA \\
        MIT Spring 2018 & 44.35 & 60.48 & 50.00 & 60.48 & 38.71 & \textbf{70.97} & 32.26 & 67.74 & 33.33 \\
        MIT Fall 2018 &	51.99 & 62.96 & \textbf{72.50} & 52.18 &	61.63 &	64.17 & 49.78 & 54.00 & 47.54 \\
        MIT Spring 2019 & 43.45 & 55.65 & \textbf{62.14} & 54.23 & 41.07 & 58.81 & 15.54 & 41.55 & 34.64 \\
        MIT Fall 2019 &	54.92 & 58.61 & 58.20 & \textbf{77.05} & 29.92 & 58.20 & 26.23 & 61.48 & NA \\
        MIT Spring 2021 & 44.33 & 48.31 & 51.26 & 55.81	& 53.45 & 60.21	& 33.62	& \textbf{62.09} & 33.77 \\
        MIT Fall 2021 &	58.94 & 61.53 & 47.24 & \textbf{69.44} & 50.35 & 54.90 & 18.11 & 42.00 & 24.44 \\
        MIT Spring 2022 &	42.78 & 55.29 & 55.07 & \textbf{68.86} & 32.03 & 53.48 & 51.01 & 65.46 &60.71 \\
        Harvard Spring 2015 &	\textbf{85.71} & 64.29 & 78.57 & NA & \textbf{85.71} & NA & 50.00 & NA & 21.43 \\
        Harvard Spring 2021 &	47.73 & 77.27 & 72.73 & \textbf{86.36} & 47.73 & 81.82 & 43.18 & \textbf{86.36} & 45.45 \\
        Cornell Spring 2017 &	78.91 & 59.90 & 79.03 & NA & \textbf{80.86} & NA & 51.30 & NA & 21.88 \\
        Cornell Fall 2018 &	36.45 & 57.01 & \textbf{61.21} & 53.27 & 44.39 & \textbf{61.21} & 42.52 & 56.07 & 28.97 \\
        \bottomrule
    \end{tabular}
\end{table*}

\begin{table*}[h!]
\small
    \caption{We benchmark different baselines for each course topic, excluding question parts that rely on images. We compare the grade of GPT-3 (text-davinci-002 and 003) with zero-shot (ZS), ChatGPT with zero-shot (ZS), GPT-3 with few-shot (FS) learning, GPT-3 with zero-shot and chain-of-thought (CoT) prompting, GPT-3 with FS and CoT, Codex with zero-shot, Codex with few-shot learning, and OPT with ZS. The question parts on loss functions rely on image information and are therefore unavailable (marked NA). The result of the best-performing method for each semester is marked in bold.}
    \label{tab:performance-by-topic}
    \centering
    \begin{tabular}{lccccccccc}
        \toprule
        Topic &	GPT-3 2 ZS & GPT-3 3 ZS & ChatGPT ZS & GPT-3 FS & GPT-3 ZS CoT & GPT-3 FS CoT & Codex ZS & Codex FS & OPT ZS\\
        \midrule
        Regression & 31.71 & \textbf{56.67} & 50.56 & 50.00 & 25.61 & 40.85 &	40.24 & 50.00 & 50.00\\
        Classifiers & 38.18 & 47.12 & 52.81 & 46.21 & 26.28 & 42.35 &	18.88 & \textbf{53.74} & 50.00\\
        Logistic Reg. & 50.00 & 60.71 & 67.86 & 60.00 & \textbf{77.50} & \textbf{77.50} & 55.00 & 70.00 & 16.67 \\
        Features & 58.65 & 71.92 & 76.15 & 75.96 & 53.85 & 77.31 & 68.85 & \textbf{81.54} & 10.00\\
        Neural Networks & 48.34 & 60.82 & 67.71 & 60.23 & 44.54 & \textbf{68.42} & 37.82 & 63.45 & 27.27\\
        CNNs & 37.50 & 59.59 & \textbf{62.05} & 53.58 & 28.36 & 47.81 & 13.38 & 36.77 & 23.83\\
        MDPs & 49.19 & \textbf{73.28} & 47.33 & 52.01 & 46.03 & 54.23 & 24.38 & 38.03 & 28.32\\
        RNNs & 61.46 & 33.33 & 45.83 & \textbf{71.88} & 57.29 & 66.14 & 12.50 & 40.63 & 39.28\\
        RL & 36.09 & \textbf{65.24} & 55.59 & 42.99 & 36.67 & 50.11 & 28.79 & 45.11 & 24.28\\
        Clustering & \textbf{100.00} & 50.00 & 90.00 & \textbf{100.00} & \textbf{100.00} & \textbf{100.00} & 50.00 & 50.00 & 63.33\\
        Decision Trees & 54.70 & \textbf{74.78} & 69.08 & 71.80 & 32.48 & 51.28 & 46.15 & 54.70 & 55.00 \\
        Model Selection & 82.93 & 71.12 & 82.10 & 83.74 & 72.76 & \textbf{95.12} & 67.48 & 69.92 & 21.95 \\
        Ensemble Methods & 27.89 & 41.35 & \textbf{69.23} & 50.00 & 22.12 & 66.35 & 32.69 & 50.00 & 13.46 \\
        Bayesian Networks & \textbf{100.00} & 0.00 & \textbf{100.00} & \textbf{100.00} & \textbf{100.00} & \textbf{100.00} & 0.00 & 0.00 & \textbf{100.00} \\
        HMMs & \textbf{100.00} & \textbf{100.00} & \textbf{100.00} & \textbf{100.00} & 50.00 & \textbf{100.00} & \textbf{100.00} & \textbf{100.00} & \textbf{100.00} \\
        Optimization & 55.00 & 77.50 & \textbf{87.50} & 60.00 & 35.00 & 55.00 & 17.50 & 70.00 & 20.00 \\
        \bottomrule
    \end{tabular}
\end{table*}

\subsection{Limitations}
\label{sec:limitations}
Our dataset consists of all question parts and their solutions, including images. However, our baseline methods do not handle questions that rely on an image containing the information required to solve the question since GPT-3, ChatGPT, and Codex do not handle images. Tables \ref{tab:image-parts-by-semester} and \ref{tab:image-parts-by-topic} show the breakdown of the number of question parts and points of questions that do not rely on image information for answering the question. On average, 27.55\% of the question parts, which make up 30.32\% of the points in final exams, are questions that rely on image information. The points attributed to the non-image parts are tallied, recorded, and used to calculate non-image percentage grades.

\begin{table}[tbh!]
\small
    \caption{The number of question parts that do not rely on images, the number of points that do not rely on images, and the number of non-open question points, in finals for each semester.}
    \label{tab:image-parts-by-semester}
    \centering
    \begin{tabular}{lccc}
        \toprule
        Semester & Non-Image & Non-Image Points & Non-Open\\
        \midrule
        MIT Fall 2017   & 49 / 61 & 70 / 100 & 69 / 70\\
        MIT Spring 2018   & 27 / 42 & 62 / 100 & 59 / 62\\
        MIT Fall 2018   & 30 / 60 & 62 / 100 & 37.5 / 62\\
        MIT Spring 2019   & 41 / 58 & 70 / 100 & 48 / 70\\
        MIT Fall 2019 & 46 / 61 & 61 / 100 & 50 / 61 \\
        MIT Spring 2021   & 51 / 71 & 62 / 100 & 43 / 61\\
        MIT Fall 2021   & 56 / 86 & 48 / 100 & 43 / 48\\
        MIT Spring 2022   & 46 / 59 & 68 / 100 & 53.5 / 68\\
        Harvard Spring 2015   & 8 / 12 & 70 / 90 & 35 / 70\\
        Harvard Spring 2021   & 15 / 32 & 22 / 53 & 11 / 22\\
        Cornell Spring 2017   & 48 / 48 & 128 / 128 & 74 / 128\\
        Cornell Fall 2018   & 51 / 56 & 107 / 120 & 71 / 107\\
        \midrule
        OPT Total & 418 / 585 & 759 / 1091 &  525 / 759\\
        \midrule
        Total & 468 / 646 & 830 / 1191 & 594 / 830\\
        \bottomrule 
    \end{tabular}
    \vspace{-10pt}
\end{table}

\begin{table}[h!]
\small
    \caption{The number of questions parts that do not rely on images, number of points that do not rely on images, and number of non-open question points in the finals for each topic of the course.}
    \label{tab:image-parts-by-topic}
    \centering
    \begin{tabular}{lccc}
        \toprule
        Topic & Non-Image & Non-Image Points & Non-Open\\
        \midrule
        Regression & 37 / 62 & 45 / 71 & 35.5 / 45\\
        Classifiers & 72 / 85 & 169 / 200 & 126.5 / 169\\
        Logistic Regression & 10 / 10 & 14 / 14 & 7 / 14\\
        Features & 18 / 21 & 26 / 38 & 22 / 26\\
        Loss Functions & 2 / 16 & 3 / 21 & 3 / 3\\
        Neural Networks & 66 / 87 & 112 / 153 & 100.5 / 112\\
        CNNs & 54 / 65 & 61 / 80 & 57 / 61\\
        MDPs & 38 / 77 & 39 / 121 & 32.5 / 39\\
        RNNs & 27 / 33 & 48 / 67 & 36 / 48\\
        RL & 52 / 60 & 87 / 111 & 57 / 87\\
        Clustering & 5 / 17 & 30 / 55 & 15 / 30\\
        Decision Trees & 33 / 48 & 76 / 113 & 39 / 76\\
        Model Selection & 16 / 16 & 41 / 41 & 14 / 41 \\
        Ensemble Methods & 20 / 20 & 52 / 52 & 31 / 52 \\
        Bayesian Networks & 1 / 6 & 2 / 11 & 0 / 2 \\
        HMMs & 1 / 4 & 1 / 7 & 0 / 1 \\
        Optimization & 13 / 16 & 20 / 32 & 15 / 20 \\
        Bonus/Name & 3 / 3 & 4 / 4 & 3 / 3 \\
        \midrule
        Total & 468 / 646 & 830 / 1191 & 594 / 830\\
        \bottomrule
    \end{tabular}
    \vspace{-10pt}
\end{table}

\subsection{Generating New Questions}
\label{sec:generation}
Creating new, high-quality questions by course instructors and TAs is crucial to assessing student learning. However, it can often be a demanding, time-consuming process that requires significant effort from those involved. These new questions must vary from past questions while testing the same core concepts. We evaluate the potential of using language models with few-shot learning to write exam content accurately and efficiently. 

\begin{figure*}[bt!]
\caption{Example of a question generated from other questions. The closest question among the example questions is very different from the new generated question.}
\label{fig:questions-to-questions}
\centering
\begin{tcolorbox}[colback=blue!5!white,colframe=blue!75!black]
\textbf{Generated Question}: ``We have a data set of images, each of which is either a picture of a dog or a picture of a cat. We want to train a binary classifier to distinguish between the two classes. Which of the following is a better choice of loss function for this task:
\begin{enumerate}[label=(\alph*)]
\item The 0-1 loss, which gives a loss of 1 if the classifier predicts the wrong class and a loss of 0 otherwise.
    
\item The cross-entropy loss, which is given by -log(p) if the correct class is predicted and -log(1-p) otherwise, where p is the predicted probability of the correct class.
    
\item The hinge loss, which is given by max(0, 1-s) if the correct class is predicted and max(0, 1+s) otherwise, where s is the predicted score for the correct class."
\end{enumerate}
\tcblower
\textbf{Closest Question}: ``Consider the following 2D dataset in (x,y) format: ((1,-1), +1), ((1,1),  +1), ((1,2.5),+1), ((2,-2),-1), ((2,1),+1),((2,3),+1),((5,-1),-1),((5,-2),-1). Break ties in distance by choosing the point with smaller $x_1$ coordinate, and if still tied, by smaller $x_2$ coordinate.
Compute the leave-one-out cross validation accuracy of the 3-nearest-neighbor learning algorithm on this dataset."
\end{tcolorbox}
\end{figure*}

\subsubsection{Questions to Questions}
We use the dataset of exam questions to generate new questions automatically. We use questions from our dataset as prompts to create new high-quality questions not present in our dataset. We create a list of various questions in our curated dataset and use the resulting list to prompt the model to create a new question.

\subsubsection{Notes to Questions}
We use the course notes to generate new programming questions automatically. We first pre-process the course notes to extract key concepts and information from the text, and then append this information to the prompt to create new high-quality programming questions. This automatic, scalable approach ensures that the generated questions closely align with the course material, which is particularly important in educational settings where the quality and relevance of questions are critical for a challenging and engaging learning experience. One significant concern may be the correctness of the generated programming question and answer. To verify the accuracy of the generated questions and answers, we tested our approach on a textbook and generated 150 questions automatically. Due to the correctness of the text, we found that all of the questions and answers were correct and are subsequently used by educators to assess students' understanding of programming concepts.

\begin{figure*}[hbt!]
\caption{Example of a programming question generated from course notes.}
\label{fig:notes-to-questions}
\centering
\begin{tcolorbox}[colback=blue!5!white,colframe=blue!75!black]
  \textbf{Prompt}: Write a medium-level programming question and answer based on: ``Bootstrap aggregation is a technique for reducing the estimation error of a non-linear predictor, or one that is adaptive to the data.
    \begin{itemize}
    \item Construct $B$ new data sets of size $n$ by sampling them with replacement from $\mathcal{D}$
    \item Train a predictor on each one: $\hat{f}^b$
    \item \textit{Regression Case}: bagged predictor is
    $$
    \hat{f}_{\text {bag }}(x)=\frac{1}{B} \sum_{b=1}^B \hat{f}^b(x)
    $$
    \item \textit{Classification Case}: majority bagged predictor: let $\hat{f}^b(x)$ be a "one-hot" vector with a single 1 and $\mathrm{K}-1$ zeros, so that $\hat{y}^b(x)=\arg \max _k \hat{f}^b(x)_k$. Then
    $$
    \hat{f}_{\text {bag }}(x)=\frac{1}{B} \sum_{b=1}^B \hat{f}^b(x),
    $$
    which is a vector containing the proportion of classifiers that predicted each class $\mathrm{k}$ for input $x$; and the predicted output is
    $$
    \hat{y}_{\text {bag }}(x)=\arg \max _{\mathrm{k}} \hat{f}_{\text {bag }}(x)_k\;\;."
    $$
    \end{itemize}
  \tcblower
  \textbf{Generated Question and Answer}: ``Implement a function in Python using the bootstrap aggregation algorithm for regression and classification. The function should take the following parameters as inputs:
    \begin{itemize}
    \item \texttt{\detokenize{dataset}}: a dataset $\mathcal{D}$ of size $n$.
    \item \texttt{\detokenize{n_models}}: an integer representing the number of models to be trained $B$.
    \item \texttt{\detokenize{train_func}}: a function that trains a single model given the training data.
    \item \texttt{\detokenize{predict_func}}: a function that takes a trained model and makes predictions on a test dataset.
    \end{itemize}
    The function should return the weighted average of all models' predictions for regression and the majority vote of all models' predictions for classification."
    
 \begin{lstlisting}
import numpy as np

def bagging(dataset, n_models, train_func, predict_func):
    n = len(dataset)
    predictions = np.zeros((n, n_models))
    for i in range(n_models):
        bootstrapped_data = np.random.choice(dataset, size=n, replace=True)
        model = train_func(bootstrapped_data)
        predictions[:, i] = predict_func(model, dataset)
    return np.mean(predictions, axis=1)

def regression_bagging(dataset, n_models, train_func, predict_func):
    return bagging(dataset, n_models, train_func, predict_func)

def classification_bagging(dataset, n_models, train_func, predict_func):
    return np.argmax(bagging(dataset, n_models, train_func, predict_func), axis=1)
  \end{lstlisting}
\end{tcolorbox}
\end{figure*}

Figures \ref{fig:questions-to-questions} and \ref{fig:notes-to-questions} demonstrate the results of this process. The Appendix consists of new generated questions from other questions and the closest question from our dataset as measured by the cosine similarity of the embedding of each question, as well as new generated questions from course notes. These new questions are diverse and qualitatively similar to questions on previous MIT final exams. This provides an efficient way for course TAs and instructors to generate new final exam questions.

\subsection{Meta-Questions}
There have been concerns about using language models in the classroom and about the models' correctness, completeness, and originality. Instead of banning usage in class, we embrace them by forming new meta-questions regarding correctness and completeness. 
Specifically, in addition to standard questions, 
we suggest a new class of final questions that consists of questions and their answers by the language model.
The students are asked to identify whether the answer is correct for each question.
If the answer is correct, then the students are asked to explain why.
If the answer is incorrect, then the students are asked to write the correct answer and provide a complete explanation. By teaching students how to use language models in an educational setting, they develop the skills they need to navigate and critically evaluate the material. Learning to prompt language models also helps students complete tasks more quickly and accurately, increasing their productivity and efficiency.

\subsection{Student Survey}
To evaluate the machine-generated questions, we conducted an anonymous online student survey comparing them with the human-written questions in terms of quality, appropriateness relative to the course, and question difficulty. We surveyed 15 students who have taken the Introduction to Machine Learning course or its equivalent. The survey was optional and included informed consent, with the following description: ``We are conducting a survey to assess the quality and difficulty of automatically generated questions for an introductory machine learning course final. You will be presented with a series of questions, either human-written (taken from an actual course final exam) or machine generated, but you will not be told the source of a given question. For each question, you will be asked (a) whether you think the question is human-written or machine-generated, (b) whether the question is appropriate for the given course final, and finally (c) how you would rate the difficulty of the question. Please carefully read each question and answer to the best of your ability''.

We randomly sampled one generated question and its closest (measured by cosine similarity) original, human-written question for each of the twelve machine learning topics. Students were asked to read these 24 questions in the survey, mixed and presented randomly, and then answer three questions for each: (1) ``Is the question human-written or machine-generated?'', (2) ``Is the question appropriate or not appropriate for the specific course final?'', and (3) ``What is the question's difficulty level on a scale between 1 (easiest) and 5 (hardest)?''. We ask the students to provide ratings and not to solve the questions. The results of our survey are as follows:
Out of the human-written questions, students identified 56.11\% of them correctly as human-written and 43.89\% incorrectly as machine-generated. Of the machine-generated questions, students identified 45\% of them correctly as machine-generated and 55\% of them incorrectly as human-written.
The difficulty ratings were between 1 (the easiest) and 5 (the hardest). Students rated machine-generated questions with a difficulty level of 2.55 with a 1.11 standard deviation and rated human-written questions with a difficulty level of 2.85 with a 1.12 standard deviation.
Students rated machine-generated questions as appropriate 82.6\% of the time and human-written questions as appropriate 85.0\% of the time.

The conclusions we draw from the survey are that (1) survey participants considered human-written questions to be as likely to be human-written or machine-generated, and similarly, machine-generated questions were considered equally likely to be machine-generated as human-written, (2) survey participants considered the machine-generated questions slightly easier than human-written questions, and (3) survey participants considered machine-generated questions as appropriate as human-written questions. Based on these results, we conclude that across multiple aspects, the machine-generated questions are highly similar to human-generated questions and can be adapted to generate questions for machine learning courses.

\subsection{Implementation Details}
\label{sec:implementation}
We use the latest OpenAI GPT-3, ChatGPT, and Codex models and do not re-train these models. We fix all the hyperparameters of the models so that the answers are deterministic and reproducible. Specifically, we set both top P, which controls diversity, and sampling temperature, which controls randomness, to $0$. The frequency and presence penalties are also set to $0$, and we do not halt any stop sequences. We allow diversity for generating new questions by setting the top P and temperature to $0.1$. We run Codex with an upper bound of generating programs with 1024 tokens. We use the OpenAI text-davinci-002 and code-davinci-002 engines for generating text and programs. For few-shot-learning and question generation, we use the text-similarity-babbage-001 engine to embed the questions and find the closest questions in the dataset by cosine similarity. The running time for answering or generating each question part is a few seconds.

\section{Conclusions}
\label{sec:conclusions}
We present a dataset and benchmark for answering and generating university-level final exams in machine learning. Machine performance and human performance are evaluated by the same graders and grading instructions, as well as by automatic checkers. A comparison of baselines shows that few-shot learning methods perform best across semesters and topics. A limitation of our work is that our benchmark does not consider questions that rely on images for their solution. This work may result in improving students learning for final exams, help course staff generate questions for finals, and compare levels of difficulty of exams across semesters and schools.

\bibliographystyle{plain}
\bibliography{bibliography}

\end{document}